\documentclass[a4paper, 10pt, conference]{ieeeconf}      
\IEEEoverridecommandlockouts                              
\usepackage{cite}
\usepackage{amsmath,amssymb,amsfonts}
\usepackage{algorithmic}
\usepackage{graphicx}
\usepackage{textcomp}
\usepackage{xcolor}
\usepackage{booktabs}
\usepackage{multirow}

\title{\LARGE \bf
Movement-Specific Analysis for FIM Score Classification Using Spatio-Temporal Deep Learning
}

\author{Jun Masaki$^{1}$, Ariaki Higashi$^{1}$, Naoko Shinagawa$^{1}$, Kazuhiko Hirata$^{2}$, Yuichi Kurita$^{1}$ and Akira Furui$^{1}$
\thanks{This work was supported by Council for Science, Technology and Innovation (CSTI), Cross-ministerial Strategic Innovation Promotion Program (SIP), ``Development of Foundational Technologies and Rules for Expansion of the Virtual Economy (JPJ012495)'' (Funding agency: NEDO).}
\thanks{$^{1}$J. Masaki, A. Higashi, N, Shinagawa, Y. Kurita and A. Furui are with the Graduate School of Advanced Science and Engineering,
Hiroshima University, Higashi-Hiroshima, 739-8527, Japan (email: {\tt\small junmasaki@hiroshima-u.ac.jp}).}
\thanks{$^{2}$K. Hirata is with the Division of Rehabilitation, Department of Clinical Practice and Support, Hiroshima University Hospital, Hiroshima, 734-8551, Japan.}
}
\begin{document}

\maketitle
\thispagestyle{empty}
\pagestyle{empty}

\begin{abstract}
The functional independence measure (FIM) is widely used to evaluate patients' physical independence in activities of daily living.
However, traditional FIM assessment imposes a significant burden on both patients and healthcare professionals.
To address this challenge,  we propose an automated FIM score estimation method that utilizes simple exercises different from the designated FIM assessment actions.
Our approach employs a deep neural network architecture integrating a spatial-temporal graph convolutional network (ST-GCN), bidirectional long short-term memory (BiLSTM), and an attention mechanism to estimate FIM motor item scores.
The model effectively captures long-term temporal dependencies and identifies key body-joint contributions through learned attention weights.
We evaluated our method in a study of 277 rehabilitation patients, focusing on FIM transfer and locomotion items.
Our approach successfully distinguishes between completely independent patients and those requiring assistance, achieving balanced accuracies of 70.09--78.79\% across different FIM items.
Additionally, our analysis reveals specific movement patterns that serve as reliable predictors for particular FIM evaluation items.
\end{abstract}

\section{Introduction}
Activities of daily living (ADL)~\cite{ADL_lawton1969} refer to the ability to independently perform basic daily activities.
Japan's rapidly aging population has led to an increased demand for rehabilitation medicine focused on maintaining and improving ADL capabilities.
For rehabilitation to be effective, accurate assessment of a patient's ADL and development of personalized treatment plans are essential.

The functional independence measure (FIM)~\cite{uds1990guide} stands out among ADL assessment methods as a widely adopted and highly reliable assessment tool.
The FIM comprises 18 items (13 motor and 5 cognitive items) rated on a seven-level scale ranging from 1 (total assistance required) to 7 (complete independence).
This comprehensive assessment enables healthcare providers to obtain detailed, objective measurements of patients' independence levels and their caregiving requirements.
However, implementing FIM requires evaluators to directly observe and score each patient's movements, imposing an increased workload on healthcare professionals and physical and psychological burden on patients.
Therefore, a technical solution to streamline the assessment process is needed.

This study aims to reduce the burden on both patients and evaluators through automated FIM score estimation from simple actions.
Rather than analyzing the specific actions defined in FIM assessment items, we propose a method that combines skeletal estimation with deep learning to estimate ADL capabilities from simple actions.
The proposed method processes three-dimensional (3D) skeletal coordinates through a deep neural network integrating a spatial-temporal graph convolutional network (ST-GCN)~\cite{ST-GCN}, a bidirectional long-short term memory (BiLSTM)~\cite{GRAVES2005602bilstm}, and an attention mechanism. 
Through estimation experiments using single actions, we investigate the relationship between simple actions and FIM motor items, working toward identifying optimal action sets for each FIM evaluation item.

\section{Related Work}
\begin{figure*}
    \centering
    \includegraphics[width=0.95\linewidth]{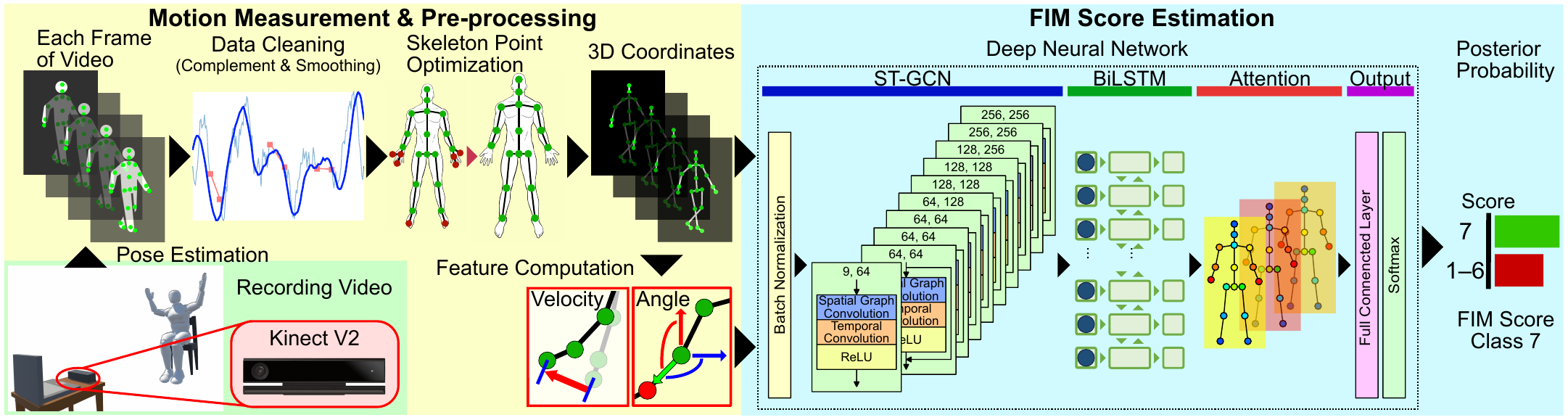}
    \caption{Overview of the proposed method. The framework integrates motion measurement, pre-processing, and deep learning-based FIM score estimation.}
    \label{fig:Overview}
\end{figure*}

\subsection{Assessment of Rehabilitation Exercise with Pose Estimation}
Recent advances in computer vision and machine learning have led to significant progress in technologies for automated human motion analysis.
In particular, skeletal estimation techniques using deep learning have enabled automatic extraction of motion features from videos simply by recording human actions.
Compared to direct video feature extraction, utilizing skeletal data allows for motion analysis that is independent of environmental factors such as recording location. 

These technologies have been successfully applied in various fields including sports science and rehabilitation support~\cite{SARDARI2023106835, deb2022gcnskl, du2021embcRehab}. 
Notably, Deb \textit{et al.}~\cite{deb2022gcnskl} addressed the challenge of evaluating whether patients correctly perform prescribed physical therapy exercises at home, which is often difficult for specialists to assess. 
They demonstrated the effectiveness of automated rehabilitation exercise assessment using pose estimation and deep learning, suggesting the potential for similar approaches in ADL assessment.

\subsection{FIM Estimation with Machine Learning}
Recent studies have explored machine learning approaches for automated FIM assessment~\cite{Matsuura2023, Oishi2019}. 
For example, Matsuura \textit{et al.}~\cite{Matsuura2023} proposed a method where participants perform several simple movements with markers attached to their bodies.
Their approach calculates multiple features from recorded motion data to estimate FIM scores using machine learning. 
This study demonstrated that FIM scores could be estimated using features derived from simple movements that differ from standard FIM evaluation items.

However, existing approaches face a main challenge: the necessity of selecting appropriate features for FIM assessment, with estimation performance heavily dependent on feature selection.
Our study addresses the limitation by developing a marker-less video-based assessment system that automatically extracts relevant motion features through deep learning, potentially enabling more efficient FIM assessment.

\section{Proposed Method}

An overview of the proposed method is shown in Fig.~\ref{fig:Overview}.
The methodology comprises three primary components: motion measurement, pre-processing, and FIM score estimation.
The process begins with the motion measurement component, where patient's movements are recorded and translated into 3D skeletal coordinates through pose estimation.
Subsequently, the pre-processing component addresses and minimizes false detections and noise artifacts generated during the pose estimation phase.
Finally, the FIM score estimation component evaluates and assigns scores for each assessment criterion.

\subsection{Motion Measurement}\label{AA}
\begin{figure}
    \centering
    \includegraphics[width=\linewidth]{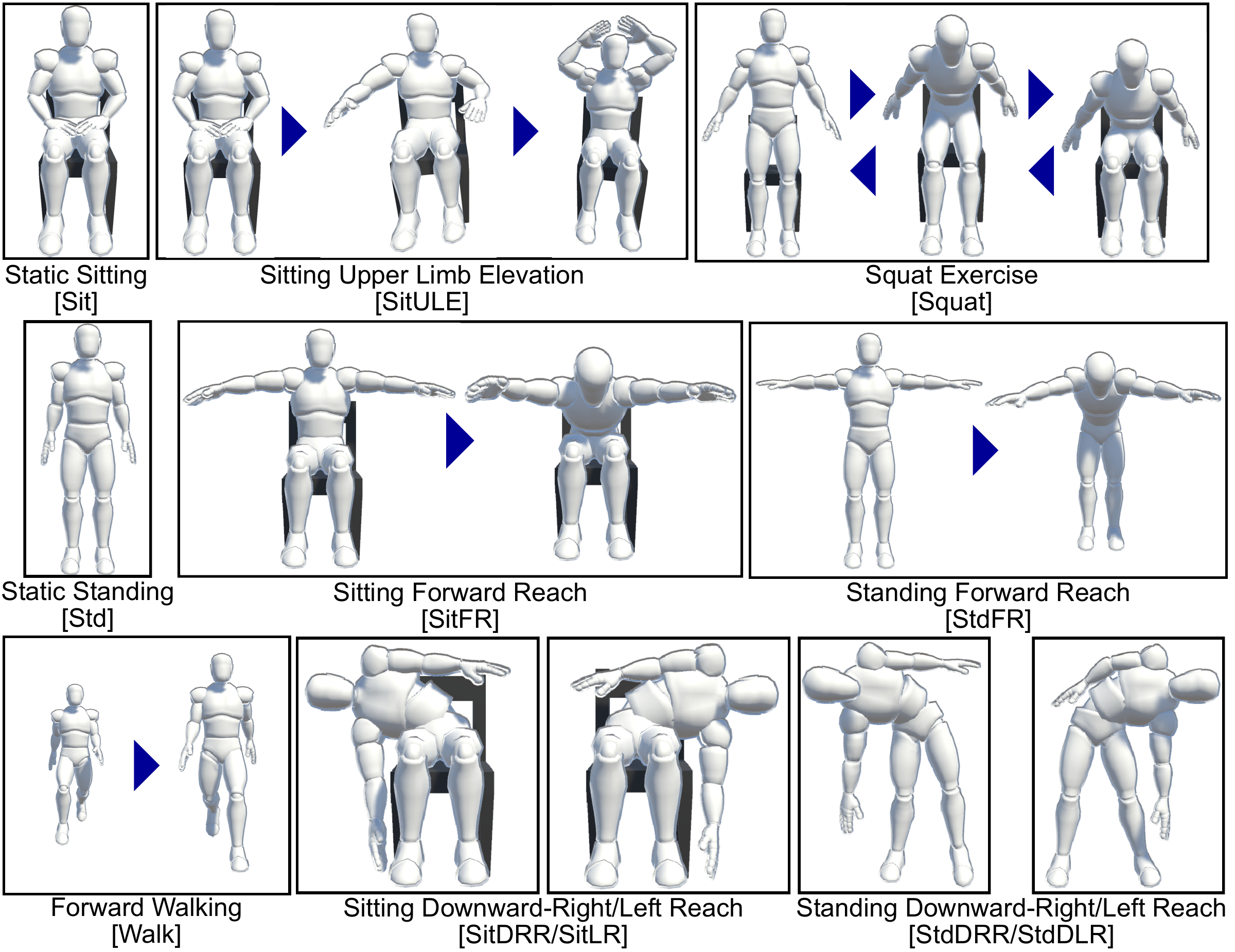}
    \caption{Measurement actions used in this study.
  Abbreviations for each action are shown in brackets.}
    \label{fig:PoseGroups}
\end{figure}

For the motion measurement component, we employ the ``AKIRA'' 
motion analysis platform (System Friend Co., Ltd.)  in conjunction with Kinect v2 (Microsoft Corporation) to conduct quantitative assessments of rehabilitation patients' movements.
The assessment protocol encompasses 11 distinct actions, as depicted in Fig.~\ref{fig:PoseGroups}: static sitting and standing postures, sitting upper limb elevation, sitting and standing forward reach, sitting and standing downward reach (both left and right), squat, and forward walking.
These specific actions were selected based on the Stroke Impairment Assessment Set (SIAS) criteria~\cite{sais1994chino}, chosen for their optimal visibility from a frontal perspective during pose estimation and their effectiveness in demonstrating arm and trunk motor functions central to SIAS assessment.

The measurement protocol involves recording approximately 10-second video segments for each movement pattern.
Pose estimation is then applied to these recordings to extract 3D coordinates of key skeletal points. 
The AKIRA system captures and stores both the video data and pose estimation results at 30 fps, integrating them with clinical data including FIM scores to create a comprehensive dataset.

\subsection{Pre-processing}
\subsubsection{Data Cleaning}
The data cleaning process addresses partial loss of body features and noise artifacts through a two-stage approach.
Linear interpolation is first applied to sections with missing frames, which restores skeletal information continuity and maintains consistent frame intervals throughout the dataset.
To reduce noise and outliers generated during pose estimation, we then apply a median filter followed by a moving average filter with a window size of $T_{\mathrm{window}}$ frames.
In these filtering operations, each filter assigns its respective statistical measure to the central frame of the window.

\subsubsection{Skeleton Point Optimization}
We optimize the skeletal model by eliminating eight extremity points---specifically, three points beyond each wrist (left and right) and one point at each foot tip---to mitigate both estimation noise and variations stemming from individual physical characteristics.
This modification aligns with our research focus on whole-body coordination and trunk motor functions, rather than fine extremity movements.
The inclusion of distal extremity data could potentially introduce unnecessary complexity into the feature extraction process.

\subsubsection{Feature Computation}\label{features}
\begin{figure}
    \centering
    \includegraphics[width=0.95\linewidth]{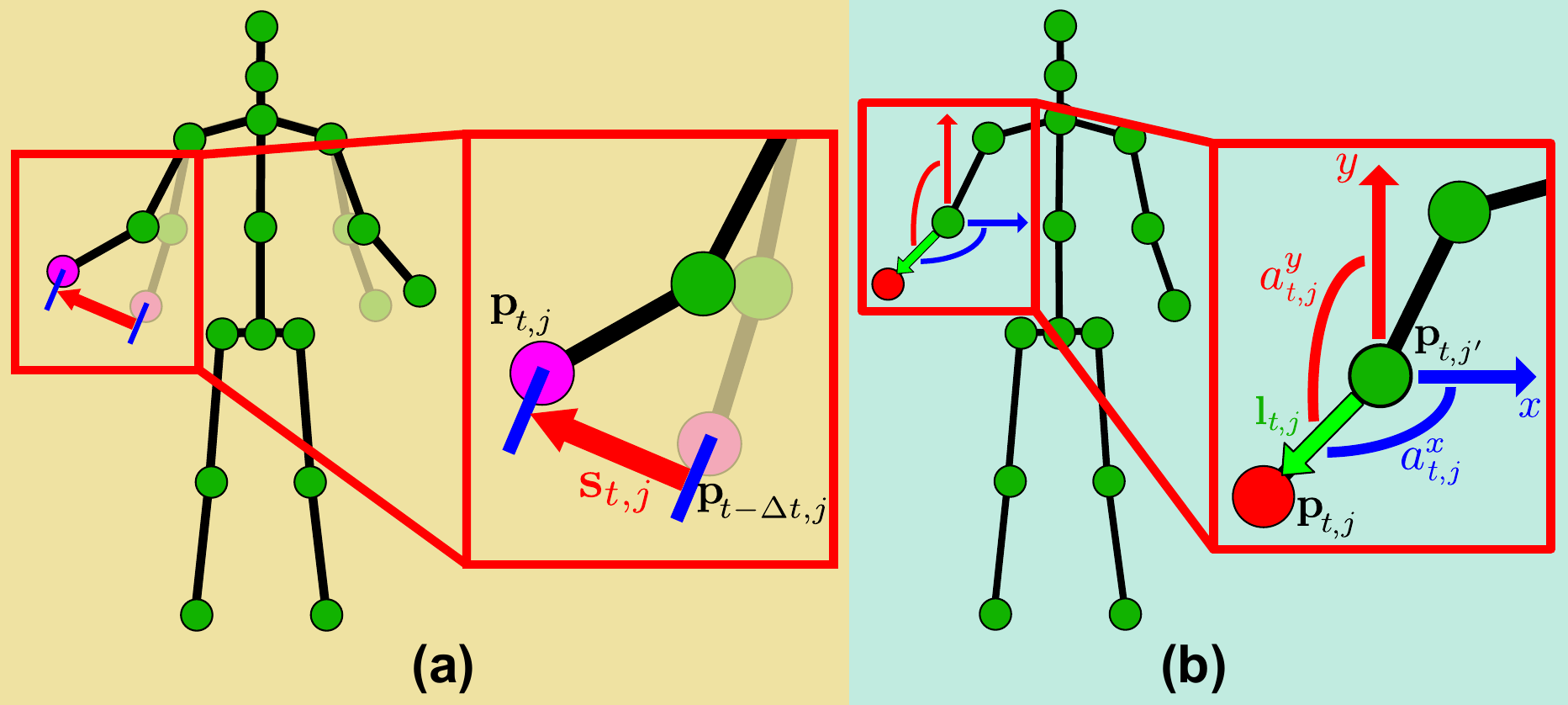}
    \caption{Illustrations showing the computation methods for (a) velocity and (b) angle features.}
    \label{fig:VelocityAngle}
\end{figure}
To effectively extract motion feature, we compute three key features for each skeletal point $j$ at frame $t$: coordinate $\mathbf{p}_{t,j} \in \mathbb{R}^3$, velocity $\mathbf{s}_{t,j} \in \mathbb{R}^3$, and the angle $\mathbf{a}_{t,j} \in \mathbb{R}^3$ between adjacent points. 
These values serve as input to the FIM score estimation component.

The velocity $\mathbf{s}_{t, j}$ of skeletal point $j$ at frame $t$ is derived from the difference between smoothed 3D coordinates of adjacent frames, normalized by the time interval (Fig.~\ref{fig:VelocityAngle}(a)):
\begin{align}
    \mathbf{s}_{t, j} = \frac{\mathbf{p}_{t, j}-\mathbf{p}_{t-\Delta t, j}}{\Delta t},
\end{align}
where $\Delta t$ represents the time interval between adjacent frames.

We compute angles using the smoothed 3D coordinates.
For each skeletal point $j$, we calculate the angle between two vectors: one pointing from $j$ to its neighboring point $j'$ (in the direction of the body center), and the other aligned with the positive direction of each coordinate axis ($x, y, z$).
The computation proceeds in two steps (Fig.~\ref{fig:VelocityAngle}(b)). 
First, we calculate the relative position vector $\mathbf{l}_{t,j}$ at frame $t$:
\begin{equation}
    \mathbf{l}_{t, j} = \mathbf{p}_{t, j} - \mathbf{p}_{t, j'}.
\end{equation}
Then, we compute the angle $a_{t,j}^w$ with respect to each axis $w \in \{x, y, z\}$:
\begin{equation}
        a_{t, j}^w = \arccos{\left(\frac{l_{t, j}^w}{\| \mathbf{l}_{t,j} \|}\right)},
\end{equation}
where $l_{t,j}^w$ represents the $w$-axis component of $\mathbf{l}_{t,j}$, and $\|\mathbf{l}_{t,j}\|$ denotes its Euclidean norm.
The angle values range from $0$ to $\pi$ radians, with skeletal points at the spine center assigned an angle of 0 radians.

\subsection{FIM Score Estimation}
The FIM score estimation module employs a deep neural network to evaluate FIM items based on single movement sequences.
The network architecture integrates a ST-GCN with a BiLSTM and an attention mechanism. 
For this study, we convert the standard seven-level FIM scores into $K$ score classes ($K \leq 7$) as prediction targets.

\subsubsection{Skeleton Graph Construction}
The foundation of our proposed method is built upon a skeletal graph structure.
We construct a spatio-temporal undirected graph $\mathcal{G} = (\mathcal{V}, \mathcal{E})$ that encompasses $J$ skeletal points across $T$ frames, capturing both spatial relationships within frames and temporal connections between frames.
The node set $\mathcal{V} = \left\{v_{t,j} \mid t = 1, 2, \ldots, T;\, j = 1, 2, \ldots, J \right\}$ comprises vertices where each $v_{t,j}$ represents skeletal point $j$ at frame $t$.
The edge set $\mathcal{E}$ defines connections between neighboring skeletal points, incorporating both anatomical adjacencies within frames and temporal links between consecutive frames for each skeletal point.

For our implementation, we combine the calculated features $\mathbf{p}_{t,j}$, $\mathbf{s}_{t,j}$, and $\mathbf{a}_{t,j}$ (described in Section \ref{features}) to form a time-series feature tensor $\mathbf{X} \in \mathbb{R}^{9 \times T \times J}$ across $J$ skeletal points. 
For each skeletal point $j$ at frame $t$, we construct the following feature vector:
\begin{equation*}
        \mathbf{x}_{t,j} = \left[p_{t, j}^{x}, p_{t, j}^{y}, p_{t, j}^{z}, s_{t, j}^{x}, s_{t, j}^{y}, s_{t, j}^{z}, a_{t, j}^{x}, a_{t, j}^{y}, a_{t, j}^{z}\right]^\top \in \mathbb{R}^9.
\end{equation*}
This comprehensive time-series feature tensor serves as input to the ST-GCN for motion feature extraction.

\subsubsection{Spatial-Temporal Graph Convolutional Network (ST-GCN)}
ST-GCN is a feedforward neural network architecture characterized by two distinct layers: a spatial graph convolution layer and a temporal convolution layer.
This structure enables the extraction of both spatial and temporal features from skeletal motion data.
We detail these components below.

\textbf{Spatial Graph Convolution:}  
This operation aggregates information from adjacent points within the skeletal graph to extract spatial movement features that reflect human body structure.
We categorize skeletal relationships into three groups based on their distance from the body center:

\begin{itemize}
    \item Root node group: connections to the node itself
    \item Centripetal group: neighboring points closer to the body center than the current node
    \item Centrifugal group: neighboring points farther from the body center than the current node
\end{itemize}

Let $f_{t, j}$ denote the edge labeling function for skeletal point $j$ at frame $t$, and $r_j$ represent the average distance from skeletal point $j$ to the body center.
For an adjacent point $v_{t,j'}$, the label $f_{t, j}(v_{t,j'})$ is defined as:
\begin{align}
f_{t, j}(v_{t, j'}) = 
\begin{cases}
    0 & \text{if } r_{j'} = r_j \text{ (Root node group)}     \\
    1 & \text{if } r_{j'} < r_j \text{ (Centripetal group)}   \\
    2 & \text{if } r_{j'} > r_j \text{ (Centrifugal group)}   
\end{cases}.
\end{align}
Based on these labels, we construct adjacency matrices $\mathbf{A}_m \in \mathbb{R}^{J \times J}$ $(m \in \{0, 1, 2\})$ where each $\mathbf{A}_m$ represents the connectivity pattern for label $m$. 

Using these adjacency matrices, the spatial graph convolution at layer $l$ is computed as:
\begin{align}
    \mathbf{F}^{(l)} = \sum_{m=0}^2 \mathbf{\Lambda}_{m}^{-\frac{1}{2}} 
    (\mathbf{A}_{m} \otimes \mathbf{M}) \mathbf{\Lambda}_{m}^{-\frac{1}{2}} \mathbf{F}^{(l-1)}\mathbf{W}_{m}^{(l)},
\end{align}
where $\mathbf{F}^{(l)} \in \mathbb{R}^{c_{l} \times \tau_{l} \times J}$ is the feature tensor at layer $l$ ($c_l$: number of output channels, $\tau_l$: number of time frames), $\mathbf{W}_{m}^{(l)} \in \mathbb{R}^{c_l \times c_{l-1}}$ is the weight matrix, $\mathbf{\Lambda}_{m} \in \mathbb{R}^{J \times J}$ is the degree matrix used for normalization, and $\mathbf{M} \in \mathbb{R}^{J \times J}$ is a learnable matrix encoding joint connection importance.
The operator $\otimes$ denotes element-wise multiplication. 

\textbf{Temporal Convolution:}  
Following spatial graph convolution, we apply a 2D convolution along the temporal axis for each skeletal point.
This step aggregates local temporal features, enabling the extraction of feature representations that incorporate both spatial structure from skeletal connections and temporal dynamics.

\subsubsection{Bidirectional Long-Short Term Memory (BiLSTM)}  
We connect a BiLSTM network after the ST-GCN to capture long-term temporal dependencies from the extracted short-term spatial features.
This architecture effectively captures temporal dependencies across the entire sequence, outputting a feature vector $\mathbf{z} \in \mathbb{R}^{C_{\mathrm{out}} \times T_{\mathrm{out}} \times J}$, where $C_{\mathrm{out}}$ and $T_{\mathrm{out}}$ represent the number of channels and the sequence length, respectively.

\subsubsection{Attention Mechanism}
To emphasize salient spatial and temporal features in FIM score estimation and enhance model interpretability, we implement an attention mechanism based on a multilayer perceptron (MLP).
This mechanism sequentially learns weights in both spatial and temporal dimensions to highlight essential information.

The attention process is formulated as follows:
\begin{align}
    \mathbf{v}_{\tau} &= \sum_{j=1}^{J} \alpha_{\tau,j} \mathbf{z}_{\tau,j}, \\ \mathbf{z}_{\mathrm{out}} &= \sum_{\tau=1}^{T_{\mathrm{out}}} \beta_{\tau} \mathbf{v}_{\tau},
\end{align}
where $\alpha_{\tau,j} \in \mathbb{R}$ and $\beta_{\tau} \in \mathbb{R}$ are attention weights for spatial and temporal dimensions, respectively.
These weights indicate the relative importance of the input feature vector $\mathbf{z}_{\tau,j} \in \mathbb{R}^{C_{\mathrm{out}}}$ and the intermediate feature representation $\mathbf{v}_{\tau} \in \mathbb{R}^{C_{\mathrm{out}}}$.
The attention weights are computed through nonlinear MLP transformations and normalized using the softmax function.

\subsubsection{Loss Function}
We train the network using a dataset of $N$ patients, $\left\{(\mathbf{X}^{(n)}, \mathbf{y}^{(n)})\right\}_{n=1}^N$, where $\mathbf{y}^{(n)} \in \{0,1\}^K$ represents the class of the ground truth FIM score encoded with one hot.
The network is optimized using the cross-entropy loss:

\begin{align}
    L = -\frac{1}{N}\sum_{n=1}^{N}\sum_{k=1}^{K}y^{(n)}_k \log{\hat{y}^{(n)}_{k}},
\end{align}

where $y^{(n)}_k$ and $\hat{y}^{(n)}_k$ denote the $k$-th elements of $\mathbf{y}^{(n)}$ and $\hat{\mathbf{y}}^{(n)}$, respectively.
The weights of all components are jointly optimized by minimizing this loss function via backpropagation through time (BPTT).

\section{Experiments}

\subsection{Dataset}
The effectiveness of the proposed method was evaluated using motion data collected from inpatients at five medical facilities in Japan: Hiroshima University Hospital, Amano Rehabilitation Hospital, Merry Hospital, Kurashiki Heisei Hospital, and Hiroshima Hiramatsu Hospital.
The study comprised 277 participants (116 males, 156 females, and 5 unspecified; mean age $71.8 \pm 14.7$ years, range 17--96 years, with 21 participants' age data unavailable).
Multiple measurement sessions were conducted for some participants.
All data collection procedures were approved by the Ethics Committee of Hiroshima University (Approval No. E-1857) and performed with informed participant consent. 
Table~\ref{tab:FIM_Score} presents the distribution of FIM scores across participants, including those with multiple measurement sessions and partial movement data.
Experienced physical therapists evaluated each FIM motor item using a standard seven-level scale.

As an initial step toward comprehensive FIM score classification, we formulated the task as a binary classification problem ($K = 2$) distinguishing between patients who can perform activities independently (FIM score 7) and those who require assistance (FIM scores 1--6).
This approach was chosen for two key reasons: it addresses the practical clinical need to identify patients requiring assistance, and it mitigates the challenge of class imbalance inherent in the original seven-level scale where lower FIM scores are typically underrepresented.
For our analysis, we selected three transfer-related and two locomotion-related items with sufficient sample sizes, excluding walking action.

\begin{table}
    \centering
    \caption{Distribution of videos of all subjects after deleting inappropriate data}
    \label{tab:FIM_Score}
    \begin{tabular}{ll|cc|c}
    \toprule
         &  & \multicolumn{3}{c}{Score distribution} \\
         \cmidrule{3-5}
        FIM category & FIM evaluation item & 1--6 & 7 & Total \\ \midrule
        Transfer & Bed, Chair, Wheelchair & 124 & 153 & 277\\
        & Toilet & 126 & 151 & 277\\
        & Tub, Shower & 161 & 115 & 276 \\ \midrule
        Locomotion &  Walk/Wheelchair & 156 & 120 & 276 \\
        & Stairs & 204 & 72 & 276 \\
    \bottomrule
    \end{tabular}
\end{table}

\subsection{Experimental Settings}
Performance evaluation was conducted using 150-frame segments extracted from the motion videos, with shorter segments excluded from analysis.
During pre-processing, both median and moving average filters were applied with a window size of $T_{\mathrm{window}} = 15$ frames.

The neural network was trained using the Adam optimizer with a batch size of 32 and a maximum learning rate of 0.01.
Training continued for 100 epochs, while the learning rate dynamically being adjusted using one-cycle learning rate scheduling. 
To address class imbalance, we implemented a weighted random sampler to ensure uniform class distribution within each mini-batch.

The dataset was partitioned using stratified holdout, with 80\% allocated for training and 20\% for testing.
Model performance was evaluated by comparing predictions against expert-annotated ground truth using balanced accuracy, which measures the average accuracy across FIM score classes while accounting for class imbalance.
The experiments were repeated with 10 different random initialization seeds, with results averaged across all runs.

\subsection{Ablation Study}
To validate the effectiveness of each component of the proposed method, we conducted the following two ablation experiments.

\subsubsection{Effectiveness of Input Features}
We evaluated the effectiveness of additional motion features by comparing two conditions: a baseline using only 3D skeletal coordinates and the proposed method using coordinates augmented with velocity and angle features. 
For both conditions, the basic ST-GCN architecture was used without BiLSTM and attention mechanism to isolate the impact of input features.

\subsubsection{Effectiveness of Adding BiLSTM and Attention Mechanism}
Using the complete feature set (coordinates, velocity, and angle), we evaluated the architectural enhancement by comparing the basic ST-GCN against our proposed network incorporating BiLSTM and attention mechanism. 
This comparison aimed to validate the effectiveness of the temporal modeling and feature weighting components.

\section{Results and Discussion}
\subsection{Ablation Study}

To validate the effectiveness of each component in our proposed method, we conducted two ablation experiments: (1) evaluating the impact of additional motion features, and (2) assessing the contribution of BiLSTM and attention mechanism.
Table~\ref{tab:table_ablation} presents the results aggregated across all actions and FIM evaluation items, showing mean and standard deviation of balanced accuracy.

First, incorporating velocity and angle features with 3D skeletal coordinates improved the balanced accuracy by 2.76\% compared to using coordinates alone.
This suggests that these additional features effectively capture temporal and topological motion characteristics that coordinate data alone cannot fully represent.

Second, the sequential addition of BiLSTM and Attention mechanism further enhanced performance, with each component contributing to improved accuracy. The complete model achieved a balanced accuracy of 67.92\%.
BiLSTM effectively captures temporal dependencies across motion sequences, and the attention mechanism emphasizes motion segments most relevant to FIM score estimation. 
Together, these architectural enhancements improved the model's ability to distinguish between different motion characteristics.

\subsection{Overall Evaluation of Proposal Method}
\begin{figure}
    \centering
    \includegraphics[width=0.95\linewidth]{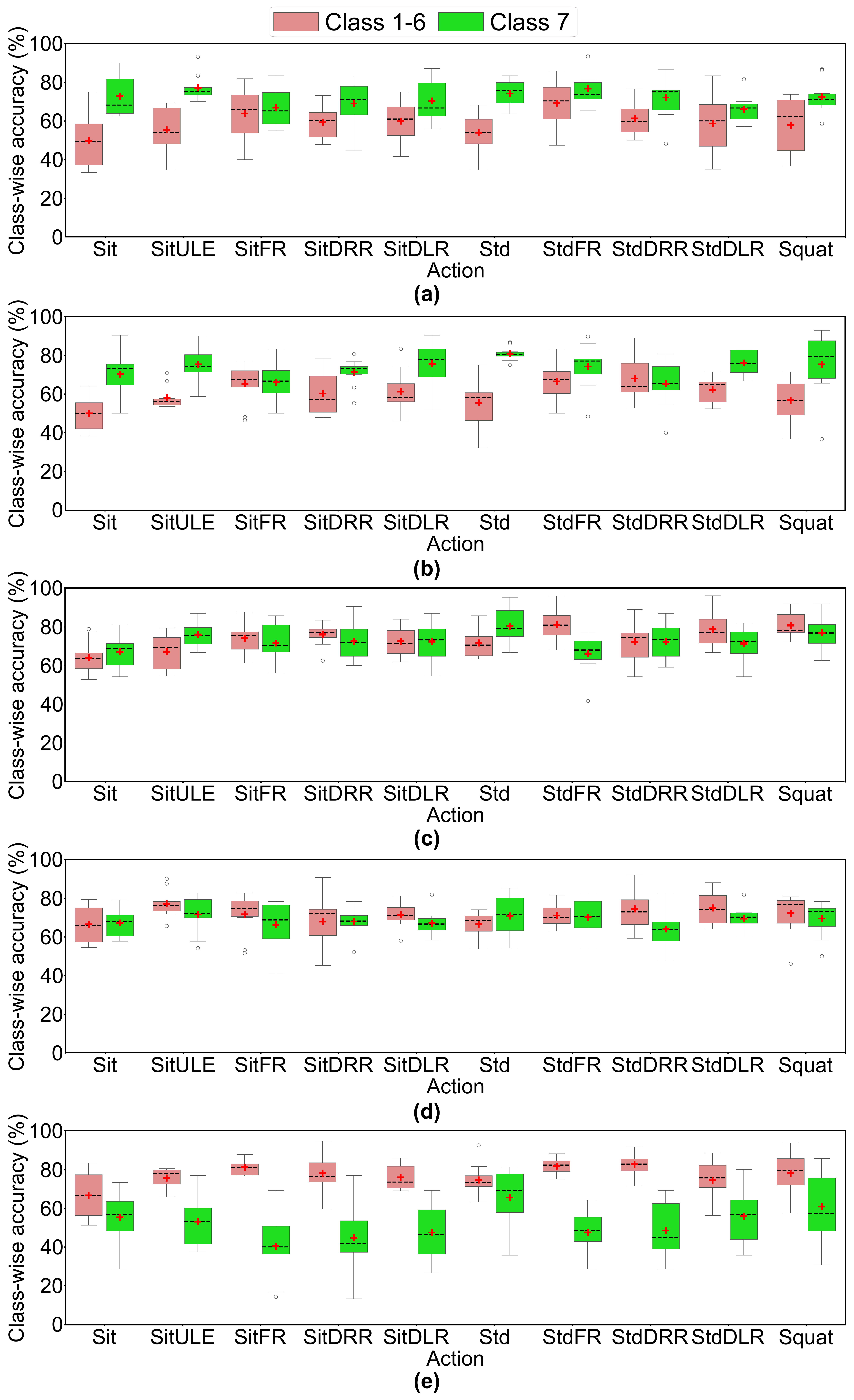}
    \caption{Class-wise accuracy distributions for each action, shown by FIM item:
  (a) Bed, Chair, Wheelchair, (b) Toilet, (c) Tub, Shower, (d)  Walk/Wheelchair, and (e) Stairs.}
    \label{fig:boxplot_accuracies}
\end{figure}

The proposed method achieved an average balanced accuracy of 67.92\% across all actions (Table~\ref{tab:table_ablation}). 
Fig.~\ref{fig:boxplot_accuracies} presents box plots showing the detailed performance distribution for each FIM evaluation item, where the horizontal axis represents input actions and the vertical axis shows class-wise accuracies. 
The median values are indicated by dashed lines, and mean values are denoted by red crosses.

\begin{table}
    \centering
    \caption{Impact of velocity/angle features and BiLSTM-attention components}
    \label{tab:table_ablation}
    \begin{tabular}{ccc|cc}
    \toprule
        Velocity/ & BiLSTM & Attention & Balanced Acc & Gain \\
        Angle & & & (\%) & (\%) \\ \midrule
         &  & & 64.82 $\pm$ 7.31 & - \\
         $\checkmark$ &  &  & 67.58 $\pm$ 7.53 & \textcolor{blue}{+2.76} \\
         $\checkmark$ & $\checkmark$ &  & 67.62 $\pm$ 7.65 & \textcolor{blue}{+2.80} \\
         $\checkmark$ & $\checkmark$ & $\checkmark$ &\textbf{67.92 $\pm$ 7.51} & \textbf{\textcolor{blue}{+3.10}} \\
         \bottomrule
    \end{tabular}
\end{table}

\subsubsection{Analysis of Each Task}
Analysis of transfer-related items revealed that SitFR and StdFR actions yielded the highest accuracies for Bed, Chair, Wheelchair, and Toilet transfers (Fig.~\ref{fig:boxplot_accuracies}(a), (b)). 
This correlation likely stems from the biomechanical similarity between these forward-reaching actions and the postural control required during chair transfer activities. 
For Tub, Shower transfers, standing reach actions (StdFR, StdDRR, StdDLR) and Squat demonstrated superior performance (Fig.~\ref{fig:boxplot_accuracies}(c)), which can be attributed to their kinematic similarity to bathtub transfer movements.

For locomotion-related items, different movement patterns emerged as predictive indicators. 
In the Walk/Wheelchair assessment, SitULE emerged as the most predictive action (Fig.~\ref{fig:boxplot_accuracies}(d)), suggesting that upper limb mobility is a key indicator of walking ability, in line with previous evidence that arm swing influences the stability of human gait~\cite{Punt2015_EffectOfArmSwing}.
For the Stairs assessment, while the overall accuracy for FIM score 7 was lower, Std and Squat actions provided relatively reliable predictions (Fig.~\ref{fig:boxplot_accuracies}(e)), indicating that whole-body postural control and lower limb strength are crucial factors in stair-climbing capability.

Table~\ref{tab:table_max_accuracy} summarizes the best-performing actions and their corresponding balanced accuracies for each FIM evaluation item.
The standard deviations reported in the table represent the variability across different random initialization seeds used in training.
The results show that specific actions can serve as effective predictors for particular FIM items, with balanced accuracies ranging from 70.09\% to 78.79\%.

\begin{figure}
    \centering
    \includegraphics[width=0.9\linewidth]{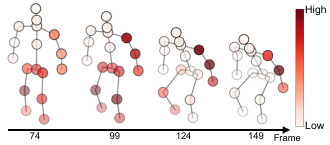}
    \caption{Example of attention weights in Squat action. The visualization shows the product of spatial ($\alpha_{\tau,j}$) and temporal ($\beta_\tau$) attention weights, where darker red indicates regions of greater focus.}
    \label{fig:attention_ex}
\end{figure}

The effectiveness of our attention mechanism is visualized through spatio-temporal joint attention weights during a Squat action (Fig.~\ref{fig:attention_ex}).
The attention patterns evolve with motion, highlighting biomechanically relevant body parts such as lower limb joints and trunk at each phase.
This temporal adaptation of attention weights demonstrates that our model successfully identifies key anatomical regions for FIM score estimation throughout the movement.

\subsubsection{Limitations}
In this study, even the best-performing actions (Table~\ref{tab:table_max_accuracy}) achieved balanced accuracies only in the 70--79\% range.
This limitation is mainly due to (i) the actions demanding fewer motor functions relative to the complex movements required by the FIM assessments and (ii) the restriction to a single action, which limits multi-faceted observation of motor capabilities.
Nonetheless, achieving over 70\% accuracy indicates that FIM scores can be inferred even from simplified, non-prescribed actions.
Classification accuracy should improve by replacing the target actions with those including more evaluation-relevant motor functions or by combining multiple actions to assess more complex motor functions comprehensively.

Although this study focused on binary classification between complete independence (FIM score 7) and the need for assistance (FIM scores 1–6), these results suggest that the method could serve as a screening tool to identify patients who may require support.
Future work should address enhancing the network architecture to achieve higher classification accuracy, enabling easy, low-burden assessment of care needs even at home.

\section{Conclusion}
\begin{table}
    \centering
    \caption{Best predictive actions and their balanced accuracy for each FIM evaluation item}
    \label{tab:table_max_accuracy}
    \begin{tabular}{l|cc}
    \toprule
        \multirow{2}{*}{} & \multicolumn{2}{c}{Best Performance} \\
        \cmidrule{2-3}
        FIM evaluation item & Action & Balanced Acc (\%) \\ 
        \midrule
        \quad Bed, Chair, Wheelchair & StdFR & 72.92 $\pm$ 5.33 \\
        \quad Toilet & StdFR & 70.33 $\pm$ 5.61 \\
        \quad Tub, Shower & Squat & 78.79 $\pm$ 5.76 \\ 
        \midrule
        \quad Walk/Wheelchair & SitULE & 74.33 $\pm$ 6.02 \\
        \quad Stairs & Std & 70.09 $\pm$ 7.23 \\
        \bottomrule
    \end{tabular}
\end{table}

In this paper, we proposed a method for automatically estimating FIM transfer and locomotion item scores from simple actions using skeletal estimation and deep learning. 
Our analysis revealed that specific actions, such as standing reach actions (StdFR, StdDRR, StdDLR), SitULE, and Squat, were particularly effective for estimating each FIM item, achieving balanced accuracies of 70.09--78.79\% when using the best predictive action for each item. 
These results demonstrate that FIM scores can be effectively estimated using actions that differ from the standard evaluation items.

In future work, we aim to develop an estimation method using multiple actions to capture features unidentifiable from a single action, while improving class imbalance handling and network architecture.

\bibliographystyle{bibstyle} 
\bibliography{references} 

\end{document}